# An entropy-based learning algorithm of Bayesian conditional trees

**Dan Geiger**
dang@cs.technion.ac.il
Computer Science Department
Technion—Israel Institute of Technology
Haifa, Israel, 32000

## Abstract

This article offers a modification of Chow and Liu's learning algorithm in the context of handwritten digit recognition. The modified algorithm directs the user to group digits into several classes consisting of digits that are hard to distinguish and then constructing an optimal conditional tree representation for each class of digits instead of for each single digit as done by Chow and Liu (1968). Advantages and extensions of the new method are discussed. Related works of Wong and Wang (1977) and Wong and Poon (1989) which offer a different entropy-based learning algorithm are shown to rest on inappropriate assumptions.

## 1 Introduction

Much work is recently being devoted to learning arbitrary Bayesian networks from data (e.g., Pearl and Verma (1991), Cooper and Herskovits (1991)). The purpose of these investigations is to uncover a Bayesian network that represents a phenomena of interest from measurements. This article develops, in the context of handwritten digit recognition, a learning algorithm of Bayesian networks that have a specific topology called a conditional tree. The proposed algorithm is very efficient and it has an entropy-based performance guarantee similar to that of Chow and Liu's (1968). The algorithm learns also Bayesian networks that have a small cutset of root nodes but its performance deteriorates as the size of the cutset increases.

The most accurate representation of a digit using just a given set of possible measurements is clearly the joint distribution $p(x, c)$ where $c$ is a digit and $x$ is a vector of measurements. However, such a representation requires exponential number of parameters which are infeasible to collect, and hard to use in computing the most likely digit. The simplest solution to this combinatorial explosion is to assume that all variables are conditionally independent given any digit. Such an assumption, however, is too restrictive and not really needed.

A more sophisticated approach was taken by Chow and Liu (1968). They devised an algorithm that selects, for each digit $c_0$, a probability distribution $\hat{p}(x \mid c_0)$ representable by a dependence tree (to be defined) that minimizes the *divergence* to $p(x \mid c_0)$ among all discrete probability distributions representable by a dependence tree. The *divergence* between $p$ and $\hat{p}$, also known as the *Kullback-Leiber measure* and *cross-entropy*, is defined as follows:

$$D(p(x \mid c_0), \hat{p}(x \mid c_0)) = \sum_x p(x \mid c_0) \cdot \log \frac{p(x \mid c_0)}{\hat{p}(x \mid c_0)}. \quad (1)$$

Information theoretic justifications for the use of this measure can be found for example in (Lewis, 1959).

Chow and Liu's algorithm turned out to be remarkably efficient; rather than examining each possible dependence tree, the algorithm computes a set of weights and identifies the maximum weight spanning tree using these weights. For each pair of variables $(x_i, x_j)$ the appropriate weight is found to be the *instantaneous mutual information* $I(x_i, x_j \mid c = c_0)$ defined by

$$I(x_i, x_j \mid c = c_0) = \sum_x p(x_i, x_j \mid c_0) \cdot \log \frac{p(x_i, x_j \mid c_0)}{p(x_i \mid c_0) \cdot p(x_j \mid c_0)}.$$

Experiments have shown that the error rate of digit recognition drops to about half compared to the error rate obtained using the naive assumption of conditional independence of all $x_i$'s given any digit (Chow and Liu, 1968).

One should notice that Chow and Liu construct several dependence trees, one for each digit, and not a single dependence tree for the entire domain. Such a collection of networks was recently termed a Bayesian multinet (Geiger and Heckerman, 1991). Chow and Liu apparently observed that the assumptions of conditional independence encoded in a single dependence tree do not represent the digits as good as the assumptions of (asymmetric) conditional independence encoded in a set of dependence trees each having a different structure that depends on the digit.

This article offers a modification of Chow and Liu's learning algorithm. The modified algorithm directs the user to group digits into several classes consisting of digits that are hard to distinguish and then constructing an optimal dependence tree representation for each class of digits instead of for each single digit as done by Chow and Liu. The algorithm finds a maximum weight spanning tree for each class of digits $D$ where the weight between two variables $x_i$ and $x_j$ is found to be the *conditional mutual information* defined by

$$\sum_{c_k \in D} p(c_k) \cdot I(x_i, x_j \mid c = c_k).$$

The main importance of this modification is that for each class of similar digits $D$, a dependence tree can now be constructed using only features that are judged (manually) to be helpful in making the distinctions between digits in $D$. Consequently, noise that is due to limited-sized samples can be removed with the help of qualitative knowledge of the digit's prototypical forms. The new learning algorithm combines an entropy-based optimization criterion together with Heckerman's similarity networks representation scheme (1991).

Another contribution made herein is an investigation into the relationship between minimization of Bayes error rate, divergence, and conditional entropy. Related works of Wong and Wang (1977) and Wong and Poon (1989) which offer a different entropy-based learning algorithm are shown to rest on inappropriate assumptions.

Although this article is self contained knowledge is assumed of the definition and usage of Bayesian networks. For details consult (Pearl, 1988), (Pearl, 1986) or (Lauritzen and Spiegelhalter, 1988).

## 2 Learning Conditional Trees

In this section, the class of functions among which an approximation for $p(x, c)$ is selected are those probability distributions representable by a *conditional dependence tree*, where $x$ consists of a finite set of variables $x_1, \ldots, x_n$ each having a finite domain and $c$'s domain is finite as well.

**Definition** A probability distribution $p(x, c)$ is representable by a *conditional dependence tree* if it is of the form:

$$\hat{p}(x, c) = p(c) \cdot \prod_{i=1}^{n} p(x_{m_i} \mid x_{a(m_i)}, c), \qquad (2)$$

where $x_{m_1}, \ldots, x_{m_n}$ is a permutation of $x_1, \ldots x_n$ and $a : \{1, \ldots, n\} \rightarrow \{0, \ldots, n\}$ is a function that identifies for each variable $x_j$ a single variable $x_{a(j)}$ (designated as the parent of $x_j$), such that $0 \leq a(j) < j$. When $a(j) = 0$, then $x_j$ has no parent other than possibly $c$.

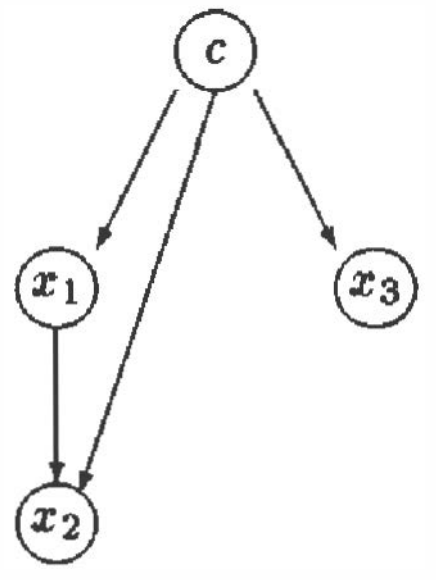

Figure 1: A conditional tree

For example $p(x_1, x_2, x_3, c) = p(c) \cdot p(x_1 \mid c) \cdot p(x_2|x_1, c) \cdot p(x_3 \mid c)$ is representable by the conditional tree shown in Figure 1. (Throughout, as a short hand notation, $x_i$ is written instead of $x_{m_i}$.)

Equivalently, a probability distribution $p$ is representable by a conditional tree if there exists a Bayesian network representation of $p$ for which node $c$ is a root node and when node $c$ is removed, the remaining network becomes a set of trees.

The probability distribution $\hat{p}(x, c)$ given by Eq. (2) is called in the literature a *product approximation* of $p(x, c)$ because it consists of a product of lower order components of $p(x, c)$ (Lewis, 1959). We return to this point in section 4.

The optimization problem can now be stated as follows. Find a probability distribution $\hat{p}(x, c)$ of the form dictated by Eq. (2) such that the divergence,

$$D(p(x, c), \hat{p}(x, c)) = \sum_{x,c} p(x, c) \cdot \log \frac{p(x, c)}{\hat{p}(x, c)}, \qquad (3)$$

is minimized.

Note the difference between Eq. (1) and Eq. (3). In the former equation which is used by Chow and Liu, a minimization procedure is exercised afresh for each digit $c_0$, while in Eq. (3), a minimization procedure is exercised once for all values of $c$ together. The motivation for this modification is clarified in the next section.

The optimization criterion is obtained by plugging the formula for $\hat{p}(x, c)$ (Eq. 2) into the divergence measure (Eq. 3):

$$D(p(x, c), \hat{p}(x, c)) =$$
$$\sum_{x,c} p(x, c) \log \frac{p(x, c)}{p(c) \cdot \prod_{i=1}^{n} p(x_i \mid x_{a(i)}, c)}$$

$$= \sum_{x,c} p(x, c) \log \frac{p(x, c)}{p(c)} - \sum_{x,c} p(x, c) \cdot$$

$$\cdot \left[ \sum_{a(i)\neq 0,\, i=1}^{n} \log \frac{p(x_i, x_{a(i)}, c)}{p(x_{a(i)}, c)} + \sum_{a(i)=0,\, i=1}^{n} \log \frac{p(x_i, c)}{p(c)} \right]$$

$$= \sum_{x,c} p(x,c) \log \frac{p(x,c)}{p(c)} - \sum_{x,c} p(x,c) \sum_{i=1}^{n} \log p(x_i \mid c)$$

$$- \sum_{x,c} p(x,c) \sum_{a(i)\neq 0,\, i=1}^{n} \log \frac{p(x_i, x_{a(i)} \mid c)}{p(x_i \mid c) \cdot p(x_{a(i)} \mid c)}$$

Note that all summands beside the last one are not affected by the choice of $a(i)$. Therefore, minimizing $D(p(x,c), \hat{p}(x,c))$ over all choices of $a(i)$ is equivalent to maximizing the summand,

$$\sum_{x,c} p(x,c) \sum_{a(i)\neq 0,\, i=1}^{n} \log \frac{p(x_i, x_{a(i)} \mid c)}{p(x_i \mid c) \cdot p(x_{a(i)} \mid c)},$$

which is equivalent to maximizing the sum,

$$\sum_{a(i)\neq 0,\, i=1}^{n} \sum_{c} p(c) \sum_{x_i, x_{a(i)}} p(x_i, x_{a_i} \mid c) \log \frac{p(x_i, x_{a(i)} \mid c)}{p(x_i \mid c) \cdot p(x_{a(i)} \mid c)}.$$

The last equation is simply the sum of conditional mutual information over all edges $(x_i, x_{a(i)})$ of the conditional tree,

$$\sum_{a(i)\neq 0,\, i=1}^{n} \sum_{c} p(c) \cdot I(x_i, x_{a(i)} \mid c). \qquad (4)$$

The optimization algorithm is now evident:

- For every pair of variables $(x_i, x_j)$ compute their mutual weight $\sum_c p(c) \cdot I(x_i, x_j \mid c)$.
- Select the maximum weight spanning tree over the $x_i$'s using known algorithms (Even, 1979) and the weights computed in the previous step.
- Add links from $c$ to each $x_i$, unless the equality $p(x_i \mid x_{a(i)}, c) = p(x_i \mid x_{a(i)})$ holds for every value of $c$.

The first two steps are justified by Eq. (4). The last step stems from the definition of a link in a Bayesian network.

## 3 Learning Networks with Small Cutset of Root Nodes

The above algorithm can be used also to learn more general Bayesian networks. Suppose $y_1, \ldots, y_l$ are root nodes and when removing them, the remaining network becomes a tree. The $y_i$'s are called a *cutset* because they break all cycles in the network. The learning algorithm for this type of networks computes the conditional mutual information between every pair $(x_i, x_j)$ conditioned on every combination of values for $y_1, \ldots, y_l$ and selects the maximum weight spanning tree over the $x_i$'s according to these weights as in the previous section. Obviously, this procedure is feasible only if the joint sample space of $y_1, \ldots, y_l$ is small enough and $y_1, \ldots, y_l$ are known in advance to be root nodes.

Rebane and Pearl (1987) extend Chow and Liu's algorithm to learn *polytrees* provided the given distributions are *polytree-isomorph*. Their algorithm can now be extended to learn any Bayesian network that has a small cutset of root-nodes. The only change needed in their algorithm is that the conditional mutual information must be computed instead of the marginal mutual information as Rebane and Pearl do. Again, this procedure is feasible only if the joint sample space of $y_1, \ldots, y_l$ is small enough and $y_1, \ldots, y_l$ are known in advance to be root nodes.

## 4 Conditional trees and Similarity networks

Section 2 develops a formula for edge weights that allows one to build a single conditional tree instead of building a tree for each digit as done by Chow and Liu. One justification for this modification is the reduction in storage obtained by replacing a set of trees with a single conditional tree. However this reduction is quite small because the total number of conditional probabilities remains unchanged. The only savings are due to the lesser overhead of storing fewer trees. The penalty in performance, on the other hand, might be quite high because if optimal trees of distinct digits are different from each other, then grouping them together yields a lesser approximation. An explicit example of this phenomena is described by Wong and Poon (1989).

The genuine motivation for the use of conditional trees is provided by Heckerman's remarkable work on Similarity networks (1991). Heckerman worked in the context of medical diagnosis but his ideas extend naturally to many pattern recognition tasks. His main observation, when translated to the terminology of digit recognition, is that not all measurements help to distinguish between every pair of digits. For example, the ratio between height and width ($r$) does not help to distinguish between a *six* and a *nine*, although it does help to distinguish between a *six* and a *one*. Therefore it is advantageous to seek a probability distribution in which the equality $p(r \mid six) = p(r \mid nine)$ is enforced, although this equality is not necessarily present in any finite sample due to noise. This is an example of qualitative knowledge that reduces the parameter space and therefore provides us with the opportunity to consider additional measurements without assuming larger samples.

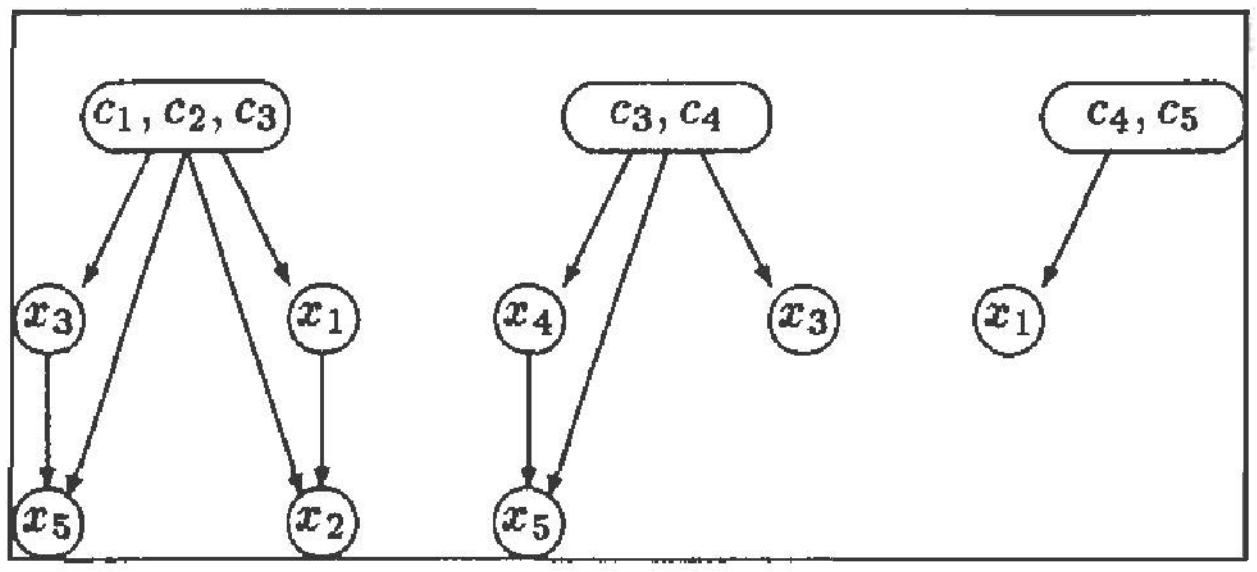

Figure 2: A similarity network

Similarity networks are defined below. A preliminary definition is needed. These definitions are borrowed from (Geiger and Heckerman, 1991).

**Definition** A *cover* of a set $A$ is a collection $\{A_1, ..., A_k\}$ of non-empty subsets of $A$ whose union is $A$. Each cover is a hypergraph, called the *similarity hypergraph*, where the $A_i$'s are called edges and elements of $A$ are called nodes. A cover is *connected* if the similarity hypergraph is connected.

For example, $\{\{0,6,9\}, \{1,7\}, \{4,7,9\}, \{5,6\}, \{2,3\}, \{3,6,8\}\}$ is a connected cover of the ten digits.

Let $\{x_1, \ldots, x_n, c\}$ be a finite set of variables each having a finite set of values, and let $p(x_1, \ldots, x_n, c)$ be a probability function having the cross product of these sets of values as its sample space. Let $A_1, ..., A_k$ be a connected cover of the values of $c$.

**Definition** A *comprehensive local network* $D_i$ of $p$ (associated with $A_i$) is a Bayesian network of the probability distribution $p(x_1 \ldots x_n | A_i)$, i.e, $D_i$ is a Bayesian network of $p(x_1, \ldots, x_n)$ given $c$ draws its values only from $A_i$. The network obtained from $D_i$ by removing nodes that are not relevant to distinguishing between elements in $A_i$ is called an *ordinary local network*. The set of $k$ ordinary local networks is called a *similarity network* of $p$.

For example, Figure 2 is an example of a similarity network representation of $p(x_1, \ldots, x_5, c)$ where $c$ has five values $c_1, \ldots, c_5$. This similarity network uses the cover $A_1 = \{c_1, c_2, c_3\}$, $A_2 = \{c_3, c_4\}$ and $A_3 = \{c_4, c_5\}$. Variable $x_1$ is the only measurement that helps to distinguish between $c_4$ and $c_5$, and variable $x_4$ is the only variable that does not help to distinguish among $\{c_1, c_2, c_3\}$. Note that many parameters of the joint distributions are not explicitly given. For example, $p(x_2 \mid x_1, c_5)$ is not given explicitly but it is equal to $p(x_2 \mid x_1, c_3)$ because $x_2$ does not appear in the local network for $\{c_3, c_4\}$ and for $\{c_4, c_5\}$.

The similarity network of Figure 2 consists of local networks each of which is a conditional tree. Each of these local networks can be constructed using the algorithm of the previous section subject to minor modifications; The weights for a link $(x_i, x_j)$ in a local network cor-

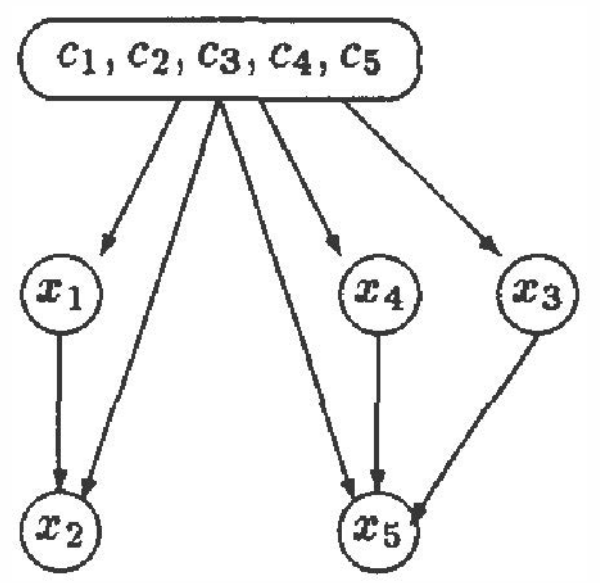

Figure 3: A global Bayesian network

responding to $A_i = \{c_{i_1}, \ldots, c_{i_l}\}$ are computed by

$$\sum_{c \in A_i} p(c) \cdot I(x_i, x_j \mid c),$$

where the sum is taken only over the values in $A_i$ instead of over all values of $c$ as done in Eq. (4), and these weights are computed only for variables $x_i, x_j$ that help to distinguish among $\{c_{i_1}, \ldots, c_{i_l}\}$.

The following procedure finds an approximation for $p(x_1, \ldots, x_n, c)$ among those probability distributions representable by a similarity network that consists of a set of conditional trees.

1. Select a connected cover $\{A_1, \ldots, A_k\}$ for the domain of $c$.
2. For each $A_i$, select the set of variables $F_i \subseteq x$ that help to distinguish between the elements in $A_i$.
3. Construct an optimal conditional tree representation for each $A_i$ using only the variables in $F_i$.
4. Apply *arc-reversal* transformations (see Shachter, 1986, 1990) until all local networks have a common ordering of nodes.
5. Combine the local networks constructed in Step 4 by taking the union of their edges (Heckerman,1990).

The first step is heuristics because it leaves the user to select any connected cover for the domain of $c$. An appropriate selection is the one that minimizes the cardinality of $\bigcup_{i=1}^{n} F_i$ where $F_i$ is defined in Step 2. Step 3 uses the algorithm of the previous section subject to the above modifications. This step is the main addition to Heckerman's work. Step 4 is required in order to prevent loops in the network produced by Step 5. Step 5 is best explained by an example; The similarity network of Figure 2 becomes, after applying Step 5, the Bayesian network shown in Figure 3. Heckerman (1990) provides the justification for Step 5. Geiger and Heckerman (1991) describe an alternative for Step 5.

The main benefit of this algorithm is attained when only a small fraction of relevant variables help to distinguish between any given pair of classes $c_i$ and $c_j$.

This condition holds in the domain of Lymph node pathology (Heckerman, 1990) where each $c_i$ denotes a disease. It seems to hold also for digit recognition where each $c_i$ denotes a prototype of some digit (the number of prototypes is about thirty). Experimental results are forthcoming.

## 5 Related optimization procedures

A simple justification for the use of the divergence measure as an approximation criterion is due to Lewis (1959). He observed that whenever $\hat{p}(x,c)$ is a product approximation of $p(x,c)$, then the divergence,

$$D(p(x,c), \hat{p}(x,c)) = \sum_{x,c} p(x,c) \cdot \log p(x \mid c) - \sum_{x,c} p(x,c) \cdot \log \hat{p}(x \mid c),$$

reduces to:

$$\sum_{x,c} p(x,c) \cdot \log p(x \mid c) - \sum_{x,c} \hat{p}(x,c) \cdot \log \hat{p}(x \mid c).$$

The first summand remains constant for all selections of $\hat{p}$. The second summand is called the conditional entropy and it can be rewritten as

$$H(\hat{p}(x \mid c)) = -\sum_{c} \hat{p}(c) \sum_{x} \hat{p}(x \mid c) \cdot \log \hat{p}(x \mid c).$$

Consequently, minimizing the divergence is a equivalent to minimizing the conditional entropy *as long as the approximate distribution is a product of lower order components of p*. This is an appealing observation because as $H(\hat{p}(x \mid c))$ is minimized, the distribution $\hat{p}(x \mid c)$ becomes sharper, and hence, on the average, each measurements' vector $x$ is more inductive about $c$. If $\hat{p}$ is not a product approximation, then the minimization of conditional entropy and cross entropy (divergence) are not necessarily equivalent.

The method of minimizing the divergence evolved as a practical substitute for the genuine optimization problem of reducing overall error rate of recognition. It is not clear that this substitute is the best feasible one to attain this goal. Wong and Wang (1977) suggest another optimization criterion: the maximization of the mutual information between $x$ and $c$,

$$I(x,c) = \sum_{x,c} \hat{p}(x,c) \cdot \log \frac{\hat{p}(x,c)}{\hat{p}(x) \cdot \hat{p}(c)}. \quad (5)$$

Wong and Wang justify their optimization criterion by showing that it is equivalent to the minimization of an upper bound of the Bayes error rate. They use an inequality between Bayes error rate $\sigma_{\hat{p}}$ and $H(\hat{p}(c \mid x))$,

$$\sigma_{\hat{p}} \leq \frac{1}{2} \cdot H(\hat{p}(c \mid x)), \quad (6)$$

obtained by Hellman and Raviv (1970) and an alternative definition for $I(x,c)$:

$$I(x,c) = H(\hat{p}(c)) - H(\hat{p}(c \mid x)) \quad (7)$$

where $H(\hat{p}(c)) = \sum_c \hat{p}(c) \cdot \log \hat{p}(c)$ and $H(\hat{p}(c \mid x)) = \sum_{x,c} \hat{p}(x,c) \cdot \log \hat{p}(c \mid x)$. Eq. (7) shows that maximizing $I(x,c)$ is equivalent to the minimization of $H(\hat{p}(c \mid x))$, and Eq. (6) shows that $H(\hat{p}(c \mid x)$ is an upper bound of Bayes error rate.

While Wong and Wang's optimization criteria is plausible its computational implementation required additional assumptions which are shown below to be inappropriate. Wong and Wang's optimization problem is stated as follows. Find a product approximation $\hat{p}$ of $p$ that minimizes $H(\hat{p}(c \mid x))$ such that, similar to Chow and Liu's work,

$$\hat{p}(x,c) = p(c) \cdot \prod_{i=1}^{n} p(x_i \mid x_{a(i)}, c), \quad (8)$$

and, contrary to Chow and Liu's work,

$$\hat{p}(x) = \sum_{c} \hat{p}(x,c) = \prod_{i=1}^{n} p(x_i \mid x_{a(i)}). \quad (9)$$

The minimization of $H(\hat{p}(c \mid x))$ is achieved by finding a maximum weight spanning tree over the $x_i$'s such that the weight for each pair $(x_i, x_j)$ is

$$\sum_{c} p(c) \cdot I(x_i, x_j \mid c) - I(x_i, x_j).$$

(Actually, this formula is Wong and Poon's version)

Notably, Wong and Wang did not notice that the assumptions encoded by Eq. (8) stand in contrast to those encoded by Eq. (9) as can be seen by the following example. Suppose only two classes $c_1$ and $c_2$ are considered and suppose $x$ consists of two binary variables $x_1$ and $x_2$. Suppose further that $x_1$ and $x_2$ are conditionally independent given $c$, i.e., for $c_1$ and for $c_2$,

$$p(x_1, x_2 \mid c_i) = p(x_1 \mid c_i) \cdot p(x_2 \mid c_i).$$

This is a common assumption which realizes Eq. (8). However, when augmented with the restriction imposed by Eq. (9),

$$p(x_1, x_2) = p(x_1) \cdot p(x_2),$$

it follows that either $x_1$ is independent of $\{c, x_2\}$ or $x_2$ is independent of $\{c, x_1\}$ (weak transitivity; Pearl, 1988). In other words, if $x_1$ and $x_2$ are conditionally independent measurements, then Eq. (9) implies that one measurement must be completely irrelevant to the classification process contrary to any reasonable interpretation.

The correct minimization problem must, therefore, ignore Eq. (9). However, minimizing

$$H(\hat{p}(c \mid x)) = H(\hat{p}(c)) - H(\hat{p}(x)) + H(\hat{p}(x \mid c)), \quad (10)$$

over all selections of $\hat{p}$ where $\hat{p}(x) = \sum_c \hat{p}(c,x)$ does not lead itself to a computationally feasible algorithm because the second summand cannot be reduced to an additive form. This complexity is perhaps the reason Wong and Wang introduced Eq. (9).

Wong and Poon (1989) undertook the task of showing that Chow and Liu's algorithm can be derived by minimizing $H(\hat{p}(c \mid x))$ instead of minimizing the divergence (Eq. 1). However, Chow and Liu's criteria is actually equivalent to minimizing $H(\hat{p}(x \mid c))$ which is different from minimizing $H(\hat{p}(c \mid x))$, as can be seen from Eq. (10). A closer look on Wong and Wang's result reveals that they assume that $H(\hat{p}(x))$ *remains constant* for all selections of $\hat{p}$ in which case the two criteria are indeed identical due to Eq. (10) and the fact that $p(c) = \hat{p}(c)$. The added assumption, however, has no mathematical or methodological justification.

## 6 Summary

A learning algorithm is presented that combines an entropy-based optimization criterion with similarity networks. This algorithm looks promising for domains of pattern recognition where a network cannot be constructed manually. A critical review of related work provides insight into the relationship between the complexity of the proposed learning algorithm viz-a-viz various entropy-based optimization criteria.

## References

[Cooper and Herskovits, 1991] Cooper, G., and Herskovits E. (1991). *Proceedings* of the 7th conference on uncertainty in artificial intelligence. Morgan Kaufmann publishers, CA.

[Chow and Liu, 1968] Chow, C. K., and Liu, C. N. 1968. Approximating discrete probability distributions with dependence trees. *IEEE Trans. on information theory*, 14(3), 462–467.

[Even, 1979] Even S. (1979). *Graph algorithms.* Computer science press.

[Geiger and Heckerman, 1991] Geiger D., and Heckerman, D. (1991). Advances in probabilistic reasoning. Proceedings of the seventh conference on uncertainty in artificial intelligence, Morgan Kaufman, 118-126.

[Heckerman, 1991] Heckerman, D. (1991). *Probabilistic similarity networks.* ACM Doctoral dissertation award series, MIT press, 1991.

[Hellman and Raviv, 1970] Hellman, M.E. and Raviv J. 1970. Probability of error, equivocation, and the Chernoff bound. *IEEE Trans. on information theory*, 16(4), 368–372.

[Lewis, 1959] Lewis, P.M. II. 1959. Approximating probability distributions to reduce storage requirements. *Information and control*, 2, 214–225.

[Lauritzen and Spiegelhalter, 1988] Lauritzen, S.L.; and Spiegelhalter, D.J. 1988. Local computations with probabilities on graphical structures and their application to expert systems (with discussion). *Journal Royal Statistical Society*, B, 50(2), 157–224.

[Pearl, 1988] Pearl, J. (1988). *Probabilistic Reasoning in intelligent systems: Networks of plausible inference.* Morgan Kaufmann, San Mateo, CA.

[Pearl, 1986] Pearl, J. (1986). Fusion, propagation, and structuring in belief networks. *Artificial Intelligence*, 29:241-288.

[Pearl and Verma, 1991] Pearl, J. and Verma, T.S. (1991). A theory of inferred causality. *Proceedings* of the second conference on the principles of knowledge representation and reasoning, Boston, MA (1991).

[Rebane and Pearl, 1987] Rebane, G. and Pearl J. (1987). The recovery of causal poly-trees from statistical data. *Uncertainty in AI 3*, eds. Kanal, Levitt, and Lemmer. Elsevier science publishers (North Holland), 1989.

[Shachter, 1986] Shachter, R. (1986). Evaluating influence diagrams. Operations Research, 34:871-882.

[Shachter, 1990] Shachter, R. (1990). An ordered examination of influence diagrams. *Networks*, 20:535-563.

[Wang and Wong, 1979] Wang, C. C., and Wong, A. K. C. 1979. Classification of discrete data with feature space transformation. *IEEE Trans. on automatic control*, 24(3), 434–437.

[Wong and Wang, 1977] Wong, A. K. C., and Wang, C. C. 1977. in *Proceedings* of the seventh conference cybernetics society, Washington, 19–21.

[Wong and Poon, 1989] Wong, S. K. M., and Poon, F. C. S. 1989. Comments on approximating discrete probability distributions with dependence trees. *IEEE Trans. on PAMI*, 11(3), 333–335.